%% file: iclr2026_conference.tex
\documentclass{article} 
\usepackage{iclr2026_conference,times}

\input{math_commands.tex}

\usepackage{hyperref}
\usepackage{url}
\usepackage{graphicx}
\usepackage{graphicx}
\usepackage{subcaption}
\usepackage{wrapfig}
\usepackage{xcolor}
\usepackage{colortbl}
\usepackage{cleveref}
\usepackage{algorithm}
\usepackage{algorithmic}
\hypersetup{
    colorlinks,
    linkcolor={red!50!black},
    citecolor={blue!50!black},
    urlcolor={blue!80!black}
}

\usepackage[most]{tcolorbox} 
\newtcolorbox[list inside=prompt,auto counter]{prompt}[1][]{
    colbacktitle=black!60,
    coltitle=white,
    fontupper=\footnotesize,
    boxsep=5pt,
    left=0pt,
    right=0pt,
    top=0pt,
    bottom=0pt,
    boxrule=1pt,
    #1,
}

\definecolor{Gray}{gray}{0.9}
\usepackage{amsmath} 
\usepackage{booktabs} 
\usepackage{booktabs}  
\usepackage{arydshln} 
\usepackage{amsmath}   
\usepackage{enumitem}

\title{Beyond Outcome Reward: Decoupling Search and Answering Improves LLM Agents}

\author{
  Yiding Wang$^{1}$ \quad Zhepei Wei$^1$ \quad Xinyu Zhu$^1$ \quad Yu Meng$^1$ \\
  $^1$Department of Computer Science, University of Virginia \\
  \texttt{blancokdb@gmail.com, \{zhepei.wei,xinyuzhu,yumeng5\}@virginia.edu}
}

\newcommand{\eg}{\textit{e.g.}}

\iclrfinalcopy 
\begin{document}

\maketitle

\begin{abstract}
Enabling large language models (LLMs) to utilize search tools offers a promising path to overcoming fundamental limitations such as knowledge cutoffs and hallucinations. 
Recent work has explored reinforcement learning (RL) for training search-augmented agents that interleave reasoning and retrieval before answering. 
These approaches usually rely on outcome-based rewards (\eg, exact match), implicitly assuming that optimizing for final answers will also yield effective intermediate search behaviors. 
Our analysis challenges this assumption: we uncover multiple systematic deficiencies in search that arise under outcome-only training and ultimately degrade final answer quality, including failure to invoke tools, invalid queries, and redundant searches. 
To address these shortcomings, we introduce \textbf{DeSA} (\textbf{De}coupling \textbf{S}earch-and-\textbf{A}nswering), a simple two-stage training framework that explicitly separates search optimization from answer generation. 
In Stage 1, agents are trained to improve search effectiveness with retrieval recall-based rewards.
In Stage 2, outcome rewards are employed to optimize final answer generation. 
Across seven QA benchmarks, DeSA-trained agents consistently improve search behaviors, delivering substantially higher search recall and answer accuracy than outcome-only baselines.
Notably, DeSA outperforms single-stage training approaches that simultaneously optimize recall and outcome rewards, underscoring the necessity of explicitly decoupling the two objectives.\footnote{Code and artifacts are available at \texttt{\href{https://github.com/yiding-w/DeSA}{https://github.com/yiding-w/DeSA}}}

\end{abstract}

\section{Introduction}

The inherent factuality limitations of large language models (LLMs), including knowledge cutoffs~\citep{chengdated} and hallucinations~\citep{huang2025survey}, have been increasingly mitigated by leveraging external tools like retrieval systems~\citep{lewis2020retrieval} and web search~\citep{wei2025webagentr1trainingwebagents}.
Early progress came from retrieval-augmented generation (RAG)~\citep{gao2023retrieval}, which grounds model outputs in relevant documents through one-shot retrieval. 
More recent advances push beyond static retrieval toward interactive search agents~\citep{jin2025searchr1trainingllmsreason,song2025r1} and ``Deep Research'' agents~\citep{zheng2025deepresearcherscalingdeepresearch,Sun2025SimpleDeepSearcherDIA} that iteratively reason, issue queries, and refine their outputs through multi-step investigation and execution.
Together, these developments mark a shift toward more autonomous, process-aware information seeking.

While early approaches like prompting~\citep{yao2023reactsynergizingreasoningacting} and supervised fine-tuning (SFT)~\citep{Song2024TrialAEA} can elicit such agentic behaviors, they face significant limitations in robustness and scalability: 
prompting alone is often sensitive to templates and generalizes poorly~\citep{sclarquantifying}, while SFT depends on costly, human-curated datasets that are difficult to scale~\citep{ouyang2022training}.
More recent work turns to reinforcement learning (RL)~\citep{kaelbling1996reinforcement,deepseekai2025deepseekr1incentivizingreasoningcapability,shao2024deepseekmathpushinglimitsmathematical,schulman2017proximalpolicyoptimizationalgorithms} as a natural framework for cultivating emergent abilities such as question decomposition, query formulation, and evidence integration. 
However, most recent RL-based methods rely almost exclusively on outcome rewards (\eg, exact match)~\citep{Fan2025SSRLSRA,jin2025searchr1trainingllmsreason}, with the underlying assumption that optimizing for final answers will indirectly teach agents to search effectively. 
This assumption is questionable: as outcome-only supervision provides sparse, delayed feedback, it may suffer from the well-known credit assignment challenges~\citep{alipov2022practicalcreditassignmentdeep,pignatelli2024surveytemporalcreditassignment}. 
In other words, it is unclear whether outcome rewards alone can effectively promote intermediate search behaviors.

\begin{figure}[!tb]
  \centering
  \includegraphics[width=.95\linewidth]{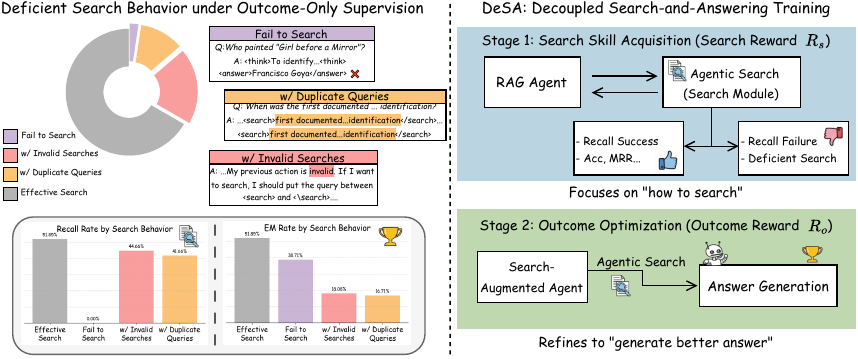} 
  \vspace{1em}
  \caption{
    An overview of deficient search behaviors of agents trained with outcome-only supervision and an illustration of our DeSA (Decoupling Search-and-Answering).
    \textbf{(Left)} The results shown are collected from an agent trained solely based on a final-answer exact match (EM) reward with Qwen2.5-3B-Instruct as the backbone, and evaluated across seven QA datasets. This agent exhibits a variety of deficient search behaviors, including ``Fail to Search'', ``w/ Invalid Searches'', and ``w/ Duplicate Queries''. Compared to ``Effective Search'', these behaviors lead to significantly lower search recall and EM rate.  
    \textbf{(Right)} DeSA decouples training into two stages to address these issues. 
  }
    \vspace{-1em}
  \label{fig:overview}
\end{figure}

To investigate this, we first conduct a systematic behavior analysis of search agents trained with outcome-only rewards~\citep{jin2025searchr1trainingllmsreason}.
In Section~\ref{sec:analysis}, we use Qwen2.5-3B/7B-Instruct~\citep{qwen2025qwen25technicalreport} as backbone models and evaluate their performance on seven question-answering benchmarks.
As exemplified in \Cref{fig:overview}, the evaluation results of Qwen2.5-3B-Instruct reveals recurring deficiencies in search behavior: (1) \textbf{Fail to Search}: agents skip retrieval altogether and rely solely on parametric memory even when external knowledge is required; (2) \textbf{w/ Duplicate Queries}: agents issue identical queries across multiple action steps, wasting budget on redundant contents; (3) \textbf{w/ Invalid Searches}: the trajectories generated by agents contain malformed tool calls or meaningless queries that yield no useful information; and (4) \textbf{Mixtures} of these issues. 
These problematic behaviors not only impair search recall and efficiency but also ultimately lead to degraded answer accuracy.

Motivated by these findings, we propose \textbf{DeSA} (\textbf{De}coupling \textbf{S}earch and \textbf{A}nswering): a two-stage RL framework that explicitly separates the processes of learning to search and learning to answer. 
In Stage 1 (Search Skill Acquisition), agents focus purely on developing effective search behaviors, guided by a retrieval recall-based reward. 
In Stage 2 (Outcome Optimization), agents leverage outcome rewards to refine their ability to distill retrieved evidence into accurate answers. By decoupling these two distinct objectives, DeSA directly addresses the limitations of outcome-only optimization. Empirical experiments show that DeSA substantially improves both search quality and QA performance. For example, DeSA reduces the deficient search rate of Qwen2.5-3B-Instruct from 23.36\% to 6.96\% and increases search recall from 59.5\% to 64.5\%. 
Beyond search metrics, DeSA also delivers superior final accuracy, surpassing strong outcome-only baselines by 8.0\% on the 3B model and 5.6\% on the 7B model across seven diverse QA benchmarks. 
These results highlight the importance of explicitly decoupling search and answering, rather than relying on outcome-only reward.

Our main contributions in this work are as follows: (1) We conduct a detailed behavioral analysis of search agents trained with outcome rewards, systematically identifying and quantifying several deficient search patterns. Our analysis reveals the insufficiency of this common training paradigm for guiding agents to develop efficient search behaviors. (2) We introduce DeSA, a decoupled two-stage RL training framework that separates \textit{Search Skill Acquisition} from \textit{Outcome Optimization}. We demonstrate that DeSA significantly improves agent performance on a wide range of QA benchmarks compared to standard single-stage outcome-based training. (3) Through extensive ablation studies, we validate that our decoupled reward design outperforms single-stage (either with outcome reward or composite reward) and other two-stage variants. 

\vspace{-10pt}
\section{Related Work}

\subsection{Retrieval-Augmented Generation (RAG)}

To address the factual hallucinations and static knowledge limitations of large language models (LLMs)~\citep{huang2025survey}, retrieval-augmented generation (RAG)~\citep{gao2023retrieval,asai2023retrieval} has emerged as an effective and widely-used technique that grounds large language models (LLMs) in external information. The concept was first introduced by \cite{lewis2020retrieval}, and since then, RAG has evolved from simple retrieve-then-generate pipelines to multi-step, iterative processes~\citep{asai2024self,Jin2024FlashRAGAMA,Gao2024ModularRTA}. These advances leverage the internal reasoning capabilities of LLMs, often elicited through techniques such as Chain-of-Thought (CoT) prompting~\citep{wei2023chainofthoughtpromptingelicitsreasoning} or post-training~\citep{Chan2024RQRAGLTA,Fang2024EnhancingNRA,wei2025instructraginstructingretrievalaugmentedgeneration}, which enable models to break down complex problems and form multi-step plans. This reframes retrieval as a sequential decision-making problem and aligns the task with the broader paradigm of tool-using agents~\citep{Tang2023ToolAlpacaGTA}, where a search engine is an invocable tool. Although prompting methods~\citep{yao2023reactsynergizingreasoningacting} guide agents through in-context examples, training-based approaches~\citep{schick2023toolformer} offer a more robust path to obtaining complex tool-use skills. However, high-quality tool-use trajectories are difficult to collect, highlighting the need for further research to improve these methods.

\subsection{Reinforcement Learning and Search Agents}

Reinforcement learning (RL)~\citep{kaelbling1996reinforcement} is a machine learning paradigm in which an agent learns to make sequential decisions by interacting with an environment and receiving feedback in the form of rewards. RL methods encompass a variety of algorithms, from Policy Gradient~\citep{schulman2017proximalpolicyoptimizationalgorithms} and Proximal Policy Optimization (PPO)~\citep{schulman2017proximalpolicyoptimizationalgorithms}, to simpler, preference-based approaches, including Direct Preference Optimization (DPO)~\citep{rafailov2024directpreferenceoptimizationlanguage} and its variants~\citep{meng2024simpo}. 
GRPO~\citep{shao2024deepseekmathpushinglimitsmathematical} improves training efficiency by replacing PPO's value model with a rule-based reward function. Building on such advancements in RL algorithms, training LLMs with RL has become a popular paradigm~\citep{zhu2025surprising,deepseekai2025deepseekr1incentivizingreasoningcapability,wei2025truthrl}. In the domain of search agents, works like Search-R1~\citep{jin2025searchr1trainingllmsreason}, DeepResearcher~\citep{zheng2025deepresearcherscalingdeepresearch}, and R1-Searcher~\citep{song2025r1} train search agents end-to-end using an outcome-based reward.~\cite{sun2025zerosearch} use simulated searches to assist in training the agent's search capabilities. Despite the extensive practical work in this field~\citep{Fan2025SSRLSRA,Jiang2025s3YDA,Sha2025SEMRLA,Sun2025SimpleDeepSearcherDIA,zhao2025parallelsearchtrainllmsdecompose}, few studies have investigated whether an outcome reward can effectively optimize search behavior.
By default, they adopt an outcome-only reward with a questionable underlying assumption that optimizing for final answers also teaches agents to search effectively. 
This motivates our investigation into decoupling search from answer generation to better optimize both tasks.

\vspace{-5pt}
\section{Preliminary}

\subsection{Search-augmented Agent}

We adopt a general search-augmented agent workflow as used in many prior works~\citep{jin2025searchr1trainingllmsreason,song2025r1,zheng2025deepresearcherscalingdeepresearch}, which is formulated as a sequential decision-making process. Given a user query $q$, the agent, equipped with an LLM policy $\pi_{\theta}$, interacts with a search engine over a series of steps to gather information before finally generating an answer. At each step $t$, the agent's state is defined by its history $H_t$, which contains the initial query and all preceding actions and observations: $H_t = (q, a_0, d_0, a_1, d_1, \dots, a_{t-1}, d_{t-1})$. Based on this history, the agent is encouraged to think and reason over retrieved information~\citep{wei2023chainofthoughtpromptingelicitsreasoning,jin2025searchr1trainingllmsreason}, and then generate the next action: $a_t \sim \pi_{\theta}(\cdot | H_t)$. The possible actions are:
\begin{itemize}[leftmargin=1em]
    \item $\texttt{search}$: Issues a search with $\texttt{query}_t$ to the search engine environment. The environment returns a set of top-ranked documents $d_t$ based on their relevance scores, which are then appended to the history. The new history becomes $H_{t+1} = H_t \oplus (a_t, d_t)$, where $\oplus$ denotes concatenation.
    \item $\texttt{answer}$: Concludes the search process and outputs the final answer to the user's question. This is a terminal action that ends the interactive process between the search agent and the environment.
\end{itemize}

The formulation highlights a strong \emph{sequential} dependence: the final answer is contingent upon the accumulated information, and each search action is chosen given the outcomes of preceding steps, highlighting the importance of search quality in the overall workflow.
Following previous works~\citep{jin2025searchr1trainingllmsreason},  we adopt a commonly used prompt template for our search agent, as shown in Table~\ref{tab:search_r1_template}.

\begin{table}[t]
\centering
\caption{Prompt template for our search agent.}
\vspace{-10pt}
\label{tab:search_r1_template}
\begin{tabular}{p{0.95\textwidth}} 
\toprule
Answer the given question. You must conduct reasoning inside \textcolor{green}{\texttt{<think>}} and \textcolor{green}{\texttt{</think>}} first every time you get new information. After reasoning, if you find you lack some knowledge, you can call a search engine by \textcolor{blue}{\texttt{<search>}} query \textcolor{blue}{\texttt{</search>}}, and it will return the top searched results between \textcolor{brown}{\texttt{<information>}} and \textcolor{brown}{\texttt{</information>}}. You can search as many times as you want. If you find no further external knowledge needed, you can directly provide the answer inside \textcolor{red}{\texttt{<answer>}} and \textcolor{red}{\texttt{</answer>}} without detailed illustrations. For example, \textcolor{red}{\texttt{<answer>}} xxx \textcolor{red}{\texttt{</answer>}}. Question: \textcolor{red}{question}. \\
\bottomrule
\end{tabular}
\end{table}
\
\vspace{-5pt}
\subsection{Reinforcement Learning Algorithms}

\textbf{Group Relative Policy Optimization (GRPO)}~\citep{shao2024deepseekmathpushinglimitsmathematical} is a reinforcement learning algorithm designed to optimize a policy without requiring an explicit value model, thereby reducing the computational and memory costs during large-scale RL training. Rather than estimating values with a separate value model, GRPO leverages a group of sampled outputs and computes a relative advantage for each response based typically on a verifiable reward function. Specifically, GRPO optimizes the following objective:
{\small
\begin{equation*}
  \mathcal{J}_{\text{GRPO}}(\theta) = \mathbb{E}_{x \sim \mathcal{D}, \{y_i\}_{i=1}^G \sim \pi_{\theta_{\text{old}}}(\cdot|x)} \left[ \frac{1}{G} \sum_{i=1}^G \frac{1}{|y_i|} \sum_{t=1}^{|y_i|} \min \left( w_{i,t}(\theta) \hat{A}_{i,t}, \text{clip}(w_{i,t}(\theta), 1-\epsilon, 1+\epsilon) \hat{A}_{i,t} \right) \right],
  \label{eq:grpo_objective}
\end{equation*}}where $G$ is the number of generated responses to each query $x$ (i.e., the group size), and the importance ratio $w_{i,t}(\theta)$ and advantage $\hat{A}_{i,t}$ of token $y_{i,t}$ are defined as:
\begin{equation*}
  w_{i,t}(\theta) = \frac{\pi_{\theta}(y_{i,t}|x, y_{i,<t})}{\pi_{\theta_{\text{old}}}(y_{i,t}|x, y_{i,<t})}, \quad \hat{A}_{i,t} = \frac{r(x, y_i) - \text{mean}\left(\{r(x, y_j)\}_{j=1}^G\right)}{\text{std}\left(\{r(x, y_j)\}_{j=1}^G\right)},
  \label{eq:grpo_advantage}
\end{equation*}
all tokens in response $y_i$ share the same advantage $\hat{A}_{i}$. Besides, GRPO directly incorporates the KL divergence term for regularization into its objective, which contributes to training stability.

\subsection{Outcome Reward}
\label{sec:em reward}
Following the standard Reinforcement Learning with Verifiable Rewards (RLVR) paradigm often employed with GRPO, previous search agent frameworks mainly use outcome reward to optimize agents' behavior. One of the representatives is Exact Match (EM) Reward:

\paragraph{Exact Match (EM) Reward}

The Exact Match (EM) reward is a binary signal that evaluates the correctness of the agent's final answer $a$. It is formally defined as:
\begin{equation*}
  R_{\text{EM}}(a, \mathcal{A}) = \begin{cases}
    1 & \text{if } \exists a^* \in \mathcal{A} \,\, \text{ s.t.\, Normalized}(a) = \text{Normalized}(a^*) \\
    0 & \text{otherwise}
  \end{cases},
  \label{eq:em_reward}
\end{equation*}
where $a$ is the agent's generated final answer, $\mathcal{A}$ is the set of ground-truth candidate answers, and $\text{Normalized}(\cdot)$ refers to a standardization function (\eg, lowercasing, removing punctuation). This reward directly supervises the accuracy of the agent's final output, making it suitable for optimizing the agent's document denoising and accurate answer generation capability.

\vspace{-5pt}
\section{Search-behavior Analysis}\label{sec:analysis}

\subsection{Deficient Search-behavior Definitions}

The central assumption of using an outcome-based reward (\eg, exact-match accuracy) for training search agents is that optimizing the final answer will implicitly guide the agent's behavior, enabling it to become an effective searcher. 
To investigate this assumption, it is necessary to systematically examine the search behaviors that arise under such training. 
In this section, we highlight several deficient behaviors that emerge when agents are trained solely with outcome rewards. 
These behaviors serve as clear indicators of failure, revealing inefficiencies and ineffectiveness in the search process that ultimately hinder the generation of accurate final answers:

\begin{itemize}[leftmargin=1em]
  \item \textbf{No Search:} The agent answers directly from its internal knowledge without using the search tool, even when the question requires external or up-to-date information. This often leads to factually incorrect or hallucinated answers.

  \item \textbf{w/ Duplicate Queries:} The agent repeatedly issues the same search query within a whole interaction. This is an inefficient strategy that wastes resources by retrieving redundant information without making progress.

  \item \textbf{w/ Invalid Searches:} The agent generates at least one invalid search action, such as mismatched or incomplete tags (\eg, \texttt{<search>query/search}), or the query is meaningless (\eg, blank space or just punctuations). Such actions result in a failed tool call without returning useful new information, wasting a step in the process.

\end{itemize}

\subsection{Analysis}

\begin{figure}[t]
\vspace{-10pt}
  \includegraphics[width=0.98\linewidth]{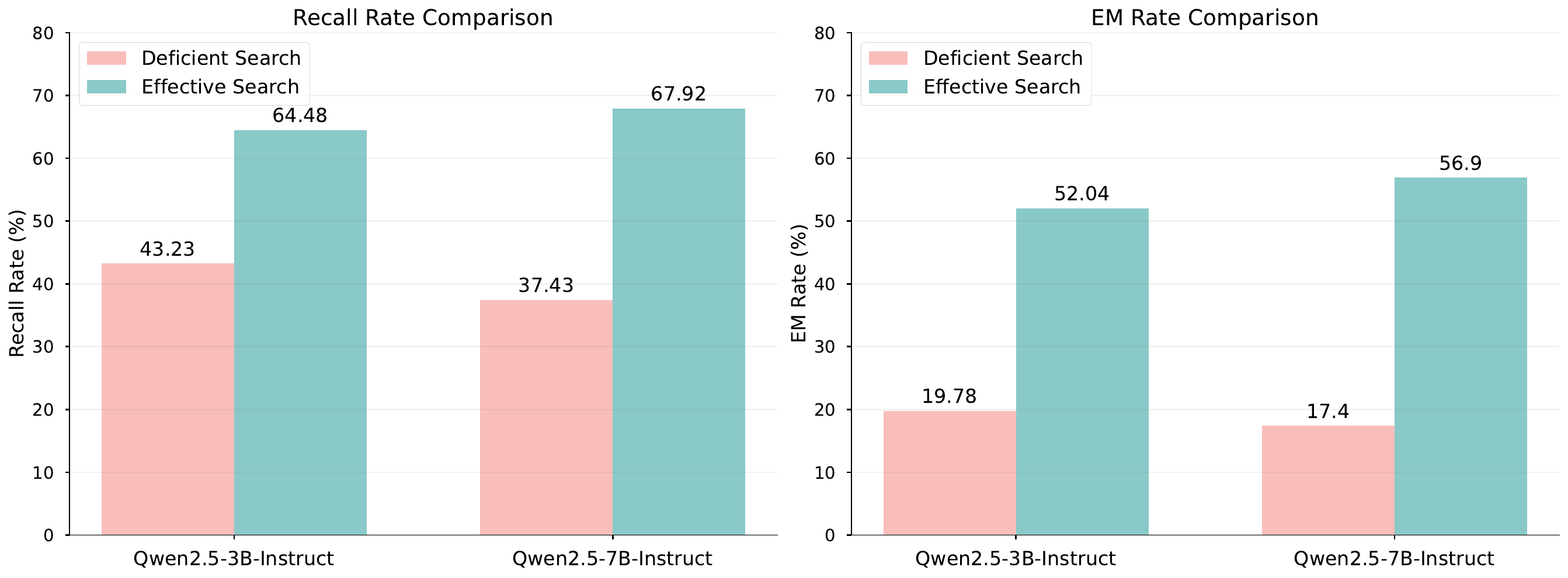}
  \vspace{-5pt}
  \caption{\textbf{Impact of deficient search behaviors on agent performance.} Both recall rate (left) and Exact Match (EM) rate (right) are significantly lower for trajectories exhibiting deficient behaviors compared to those with only effective behaviors.}
  \label{fig:performance_comparison}
  \vspace{-10pt}
\end{figure}

To investigate the effectiveness of EM reward in optimizing search behaviors, we train search agents with EM reward on NaturalQuestions~\citep{kwiatkowski-etal-2019-natural} and HotpotQA~\citep{yang2018hotpotqadatasetdiverseexplainable}, using Qwen2.5-3B-Instruct and Qwen2.5-7B-Instruct~\citep{qwen2025qwen25technicalreport} as backbones. We then evaluate their performance and inspected their search behavior on seven QA evaluation benchmarks (NaturalQuestions~\citep{kwiatkowski-etal-2019-natural}, TriviaQA~\citep{joshi-etal-2017-triviaqa}, PopQA~\citep{mallen-etal-2023-trust}, HotpotQA~\citep{yang2018hotpotqadatasetdiverseexplainable}, 2WikiMultiHopQA~\citep{ho-etal-2020-constructing}, Musique~\citep{trivedi-etal-2022-musique}, and Bamboogle~\citep{press2023measuringnarrowingcompositionalitygap}), all the metrics are aggregated over these benchmarks. We define two goals for the search agent and examine its accomplishment under different search behaviors: 1. \textbf{Recall Success}: The retrieved information contains the correct answer. 2. \textbf{Answer Correctness (EM)}: The generated answer exactly matches a reference (or ground-truth) answer.

\begin{figure}[t]
\vspace{-10pt}
  \includegraphics[width=0.98\linewidth]{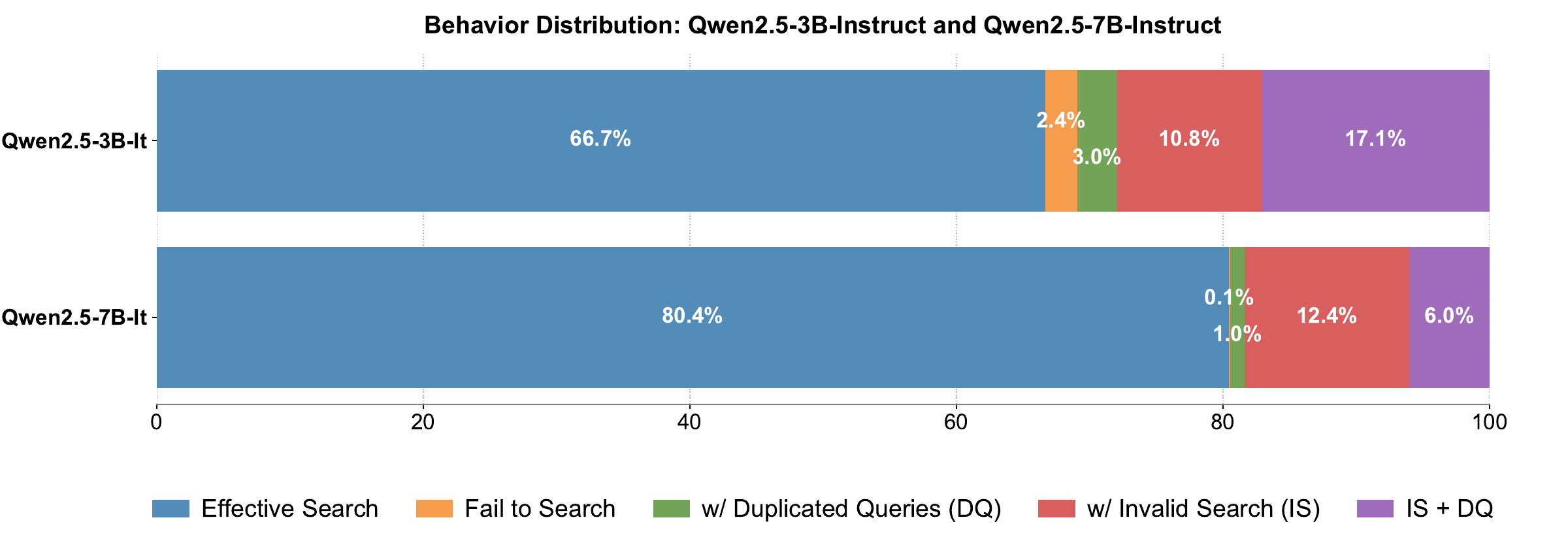}
  \vspace{-5pt}
  \caption{\textbf{Deficient Search Behaviors in Recall-Failure Cases.} This figure displays the distribution of the three defined deficient search behaviors, as well as their combinations, within all search trajectories that failed to recall the ground-truth answer.}
  \label{fig:analysis_results}
  \vspace{-10pt}
\end{figure}

To validate whether \textbf{deficient search behaviors will decrease the search agent's performance}, we first divide all interaction trajectories into two distinct groups: those containing at least one of the deficient behaviors, and those with none. A clear performance gap is observed between these two groups, as shown in Figure~\ref{fig:performance_comparison}: For the Qwen2.5-3B agent, flawed trajectories have a significantly lower average recall rate \textbf{(43.23\% vs. 64.48\%)}. This trend holds for the Qwen2.5-7B agent \textbf{(37.43\% vs 67.92\%)}. Similarly, the final EM rate on answer generation \textbf{dropped by 32.26\%} \textbf{(19.78\% vs. 52.04\%)} for the 3B model and \textbf{39.5\%} \textbf{(17.4\% vs. 56.90\%)} for the 7B model. To further dissect the nature of these failures, we analyzed the composition of all cases where the agent failed to recall the answer. As illustrated in Figure~\ref{fig:analysis_results}, our findings are: for the Qwen2.5-3B model, \textbf{33.3\%} of all recall failures exhibited at least one of the deficient behaviors. This issue holds for the larger Qwen2.5-7B model, where the proportion is 19.58\%. More seriously, the ``Effective Search'' behavior only denotes that it is clean-formed without collapsed patterns; there is still room to make it ``Efficient'' and achieve a higher recall rate. All of our analysis results provide strong evidence that \textbf{training with EM reward alone does not effectively translate the outcome reward signal into the optimization of intermediate search actions}, leading to frequent and predictable failure modes or inefficient searches.

\section{Method}

Our analysis shows that relying solely on an outcome-based EM reward does not effectively guide the intermediate search steps, leading to deficient and inefficient behaviors that reduce both recall and final EM accuracy. 
We therefore reconsider the search agent's sequentially dependent workflow and argue that successful search is a critical prerequisite for producing well-grounded answers. 
Building on this insight, we introduce DeSA (Algorithm~\ref{alg:desa}), a two-stage decoupled training framework that explicitly separates these two objectives and applies tailored supervision sequentially: first, we train the agent exclusively on search efficacy, and then we train it to generate accurate answers from the evidence it has learned to retrieve effectively. 

\subsection{Stage 1: Search Skill Acquisition}

Given that an agent cannot answer questions for which it lacks sufficient supporting evidence, our objective in the first training stage is to provide a reward based on the retrieved information. This encourages the agent to gather more useful information, enhancing its search efficiency and efficacy, and mitigating deficient and inefficient search behaviors. We use the \textbf{Recall Reward} ($R_{\text{recall}}$) as our main objective. This reward provides a direct signal indicating whether the necessary information to answer the question has been successfully retrieved.
\paragraph{Recall Reward.} Let $\mathcal{C} = \{c_1, c_2, \dots, c_k\}$ be the set of all information blocks retrieved by the agent across its $k$ interactions. We define the aggregated recalled information $I_{\text{recalled}} = \text{Aggregate}(\mathcal{C})$. The recall reward is then formally given by:
\begin{equation}
  R_{\text{recall}}(I_{\text{recalled}}, \mathcal{A}) = \begin{cases}
    1 & \text{if } \exists a^* \in \mathcal{A} \text{ s.t. } a^* \in I_{\text{recalled}} \\
    0 & \text{otherwise}
  \end{cases}
  \label{eq:recall_reward}
\end{equation}
This reward specifically incentivizes the agent to strategically generate search queries and effectively utilize the search engine to find supporting evidence, acting as a crucial signal for improving its information-seeking behavior. We also explored alternative search rewards for Stage 1 optimization. These included: (1) a composite reward that combines $R_{\text{recall}}$ with a penalty term to discourage deficient behaviors like duplicate queries. (2) a more fine-grained 'retrieval accuracy' reward ($R_{\text{acc}}$)that measures the proportion of retrieved documents containing the answer. Through experiments, we find that the simple recall reward generally works better than alternative (and more complicated) search reward designs, as discussed in Section~\ref{sec:reward design}.

\subsection{Stage 2: Outcome Optimization}
After establishing foundational search skills in Stage 1, the second stage shifts the optimization focus to the outcome. The primary goal is to train the agent to effectively translate retrieved information into correct answers by enhancing its ability to de-noise documents, synthesize evidence, and generate a precise final solution. For this purpose, we fine-tune the agent using the \textbf{outcome reward (Exact Match)} ($R_{\text{EM}}$).

By initializing this stage with the model checkpoint from Stage 1, 
we ensure that the agent builds upon its improved search capabilities while still maintaining reasonable behaviors. Stage 2 training also allows the agent to adapt its search behavior from exhaustive search to more accurate under the optimization pressure on the final answer generation task. This stage ensures the agent leverages the higher-quality context provided by its search skills and achieves superior downstream performance.

\begin{algorithm}[t]
\caption{The DeSA Training Framework}
\label{alg:desa}
\begin{algorithmic}[1]
\REQUIRE Pre-trained LLM $\mathcal{M}_{\theta}$, Search Environment $\mathcal{E}$, Training Data $\mathcal{T} = \{\langle q, \mathcal{A} \rangle\}$
\STATE \COMMENT{\textbf{Stage 1: Search Skill Acquisition}}
\FOR{each training step in Stage 1}
    \STATE Sample question-answer pair $\langle q, \mathcal{A} \rangle$ from $\mathcal{T}$
    \STATE Agent $\mathcal{M}_{\theta}$ interacts with environment $\mathcal{E}$ given $q$ to retrieve a set of information blocks $\mathcal{C}$
    \STATE Aggregate recalled information $I_{\text{recalled}} = \text{Aggregate}(\mathcal{C})$
    \STATE Compute \textbf{Search Reward} $R_{\text{recall}}(I_{\text{recalled}}, \mathcal{A})$ using Eq.~\ref{eq:recall_reward}
    \STATE Update parameters $\theta$ of agent $\mathcal{M}_{\theta}$ to maximize $R_{\text{recall}}$
\ENDFOR
\STATE
\STATE \COMMENT{\textbf{Stage 2: Outcome Optimization}}
\FOR{each training step in Stage 2}
    \STATE Sample question-answer pair $\langle q, \mathcal{A} \rangle$ from $\mathcal{T}$
    \STATE Agent $\mathcal{M}_{\theta}$ performs search and generates a final answer $a_{\text{pred}}$ for question $q$
    \STATE Compute \textbf{Outcome Reward (EM)} $R_{\text{EM}}(a_{\text{pred}}, \mathcal{A})$
    \STATE Update parameters $\theta$ of agent $\mathcal{M}_{\theta}$ to maximize $R_{\text{EM}}$
\ENDFOR
\RETURN Trained Agent $\mathcal{M}_{\theta}$
\end{algorithmic}
\end{algorithm}

\section{Experiments}


\subsection{Experimental Setup}

\paragraph{Setup} To ensure a fair and direct comparison, we follow the implementation of Search-R1~\citep{jin2025searchr1trainingllmsreason} and keep our experimental settings closely aligned with it. We utilize Qwen2.5-3B-Instruct and Qwen2.5-7B-Instruct~\citep{qwen2025qwen25technicalreport} as the backbone large language models and GRPO as the RL algorithm for our agents. We follow previous work and integrate the loss masking for the retrieved tokens. For the training, we use a training dataset includes the training splits of Natural Questions (NQ)~\citep{kwiatkowski-etal-2019-natural} and HotpotQA~\citep{yang2018hotpotqadatasetdiverseexplainable}. We use the 2018 Wikipedia as the knowledge corpus and E5~\citep{wang2022text} as the retriever, which fetches the top 3 relevant passages for each search query.

\paragraph{Datasets} We evaluate our models on a comprehensive suite of seven question-answering benchmarks to assess both in-domain and out-of-domain performance. These datasets include: (1) General QA: NaturalQuestions, TriviaQA~\citep{joshi-etal-2017-triviaqa}, and PopQA~\citep{mallen-etal-2023-trust}; and (2) Multi-Hop QA: HotpotQA, 2WikiMultiHopQA~\citep{ho-etal-2020-constructing}, Musique~\citep{trivedi-etal-2022-musique}, and Bamboogle~\citep{press2023measuringnarrowingcompositionalitygap}. Following standard practice for these benchmarks, we use Exact Match (EM) as the primary evaluation metric.

\paragraph{Baselines} Our primary baseline is the single-stage training approach from Search-R1~\citep{jin2025searchr1trainingllmsreason}, which uses only the EM reward.
For a broader comparison, we also include baselines with other existing methods (\eg, RAG, IRCoT~\citep{trivedi2023interleavingretrievalchainofthoughtreasoning}, and SFT), citing the results reported in the Search-R1 paper.

\subsection{Main Results}

\begin{table}[t]
\centering
\caption{Main results. The best performance is set in bold. $^\dagger$ / $^*$ represents in-domain/out-of-domain datasets.}
\vspace{-.5em}
\label{tab:main_results}
\resizebox{\textwidth}{!}{
\begin{tabular}{lcccccccc}
\toprule
\multicolumn{1}{c}{\textbf{Methods}} & \multicolumn{3}{c}{\textbf{General QA}} & \multicolumn{4}{c}{\textbf{Multi-Hop QA}} & \multicolumn{1}{c}{} \\
\cmidrule(lr){2-4} \cmidrule(lr){5-8}
 & NQ$^\dagger$ & TriviaQA$^*$ & PopQA$^*$ & HotpotQA$^\dagger$ & 2Wiki$^*$ & Musique$^*$ & Bamboogle$^*$ & \textbf{Avg.} \\
\midrule
\multicolumn{9}{c}{\textbf{Qwen2.5-7b-Instruct}} \\
\midrule
Direct Inference & 0.134 & 0.408 & 0.140 & 0.183 & 0.250 & 0.031 & 0.120 & 0.181 \\
CoT & 0.048 & 0.185 & 0.054 & 0.092 & 0.111 & 0.022 & 0.232 & 0.106 \\
IRCoT & 0.224 & 0.478 & 0.301 & 0.133 & 0.149 & 0.072 & 0.224 & 0.239 \\
Search-o1 & 0.151 & 0.443 & 0.131 & 0.187 & 0.176 & 0.058 & 0.296 & 0.206 \\
RAG & 0.349 & 0.585 & 0.392 & 0.299 & 0.235 & 0.058 & 0.208 & 0.304 \\
SFT & 0.318 & 0.354 & 0.121 & 0.217 & 0.259 & 0.066 & 0.112 & 0.207 \\
R1-base & 0.297 & 0.539 & 0.202 & 0.242 & 0.273 & 0.083 & 0.296 & 0.276 \\
R1-instruct & 0.270 & 0.537 & 0.199 & 0.237 & 0.292 & 0.072 & 0.293 & 0.271 \\
Rejection Sampling & 0.360 & 0.592 & 0.380 & 0.331 & 0.296 & 0.123 & 0.355 & 0.348 \\
Search-R1 & 0.429 & 0.623 & 0.427 & 0.386 & 0.346 & 0.161 & \textbf{0.400} & 0.396 \\
\rowcolor{Gray}
\textbf{DeSA (Ours)} & \textbf{0.468} & \textbf{0.631} & \textbf{0.440} & \textbf{0.424} & \textbf{0.374} & \textbf{0.197} & 0.395 & \textbf{0.418} \\
\midrule
\multicolumn{9}{c}{\textbf{Qwen2.5-3b-Instruct}} \\ 
\midrule
Direct Inference & 0.106 & 0.288 & 0.108 & 0.149 & 0.244 & 0.020 & 0.024 & 0.134 \\
CoT & 0.023 & 0.032 & 0.005 & 0.021 & 0.021 & 0.002 & 0.000 & 0.015 \\
IRCoT & 0.111 & 0.312 & 0.200 & 0.164 & 0.171 & 0.067 & 0.240 & 0.181 \\
Search-o1 & 0.238 & 0.472 & 0.262 & 0.221 & 0.218 & 0.054 & \textbf{0.320} & 0.255 \\
RAG & 0.348 & 0.544 & 0.387 & 0.255 & 0.226 & 0.047 & 0.080 & 0.270 \\
SFT & 0.249 & 0.292 & 0.104 & 0.186 & 0.248 & 0.044 & 0.112 & 0.176 \\
R1-base & 0.226 & 0.455 & 0.173 & 0.201 & 0.268 & 0.055 & 0.224 & 0.229 \\
R1-instruct & 0.210 & 0.449 & 0.171 & 0.208 & 0.275 & 0.060 & 0.192 & 0.224 \\
Rejection Sampling & 0.294 & 0.488 & 0.332 & 0.240 & 0.233 & 0.059 & 0.210 & 0.265 \\
Search-R1 & \textbf{0.397} & 0.565 & 0.391 & 0.331 & 0.310 & 0.124 & 0.232 & 0.336 \\
\rowcolor{Gray}
\textbf{DeSA (Ours)} & 0.375 & \textbf{0.575} & \textbf{0.397} & \textbf{0.352} & \textbf{0.363} & \textbf{0.134} & \textbf{0.347} & \textbf{0.363} \\
\bottomrule
\end{tabular}
} 
\end{table}

\paragraph{Overall Performance}
\begin{wrapfigure}{r}{0.4\textwidth}
  \centering 
  
  \includegraphics[width=\linewidth]{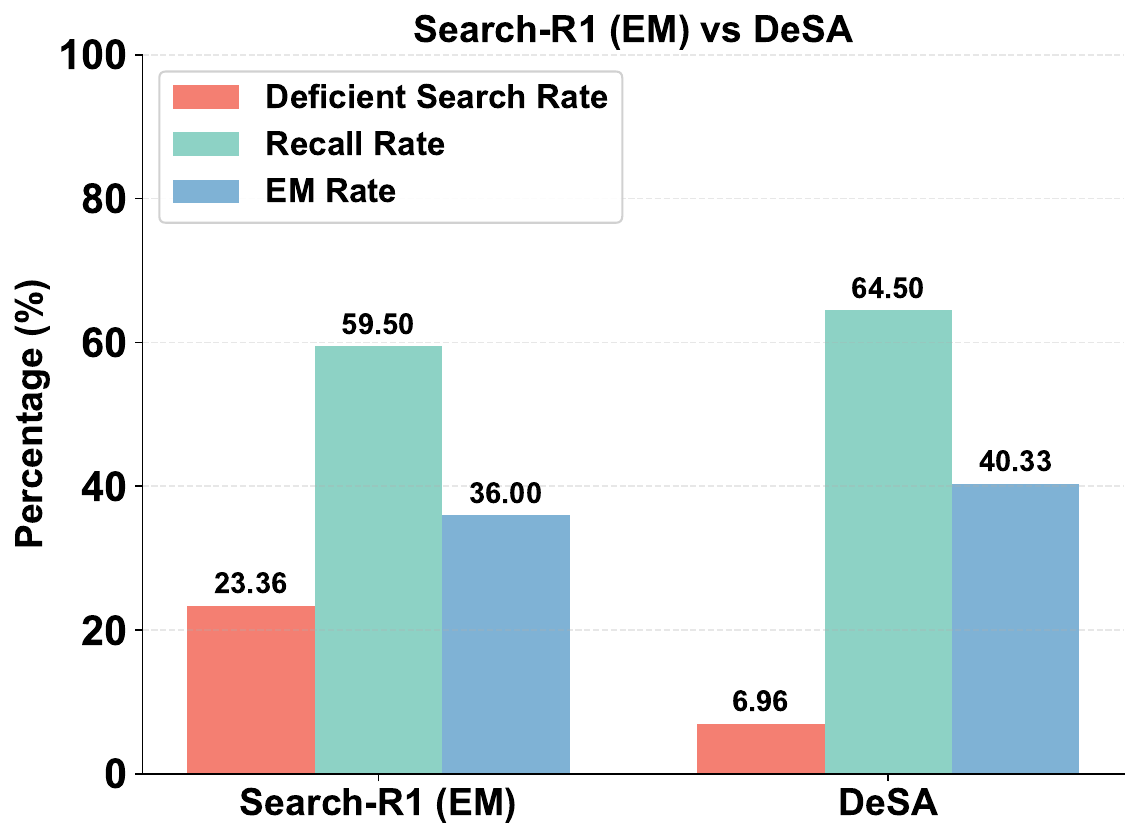}
  
  \caption{Final performance comparison of DeSA vs. the single-stage Search-R1 baseline on the 3B model.}
  \label{fig:behav_initial}
\end{wrapfigure}
As presented in Table~\ref{tab:main_results}, our proposed DeSA significantly outperforms all baseline methods across both model sizes. For the larger Qwen2.5-7B-Instruct model, DeSA achieves the top average score of 0.418, representing a 5.6\% relative improvement over the strong single-stage Search-R1 baseline (0.396). The performance gains are particularly notable on complex multi-hop question-answering tasks such as HotpotQA (+3.8 points) and Musique (+3.6 points), demonstrating DeSA's ability to facilitate more effective search strategies. The advantages of our decoupled pipeline are even more pronounced on the smaller Qwen2.5-3B-Instruct model. DeSA achieves an average score of 0.363, outperforming Search-R1 (0.336) by 8.0\%. The gains are especially significant on challenging out-of-domain datasets, including a remarkable 11.5-point absolute improvement on Bamboogle (0.347 vs. 0.232) and a 5.3-point gain on 2WikiMultiHopQA. The fact that the performance gap widens on the smaller model suggests that our two-stage approach provides a more crucial and effective learning signal, compensating for the model's reduced capacity by explicitly rewarding the acquisition of search skills before optimizing the outcome.

\begin{figure}[t]

\centering
    \includegraphics[width=\linewidth]{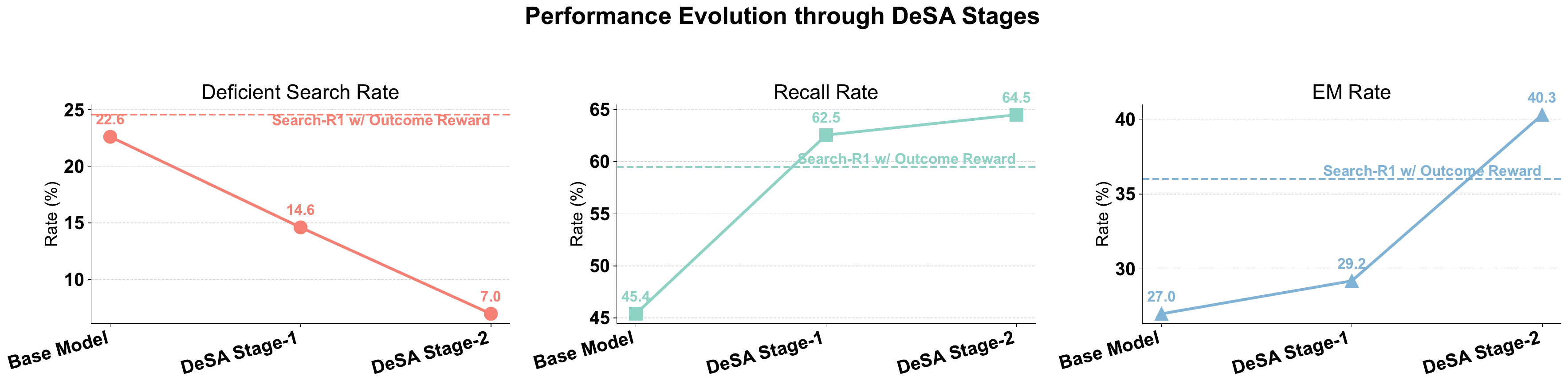}
    \vspace{-1em}
    \caption{\textbf{Performance breakdown across DeSA's two stages}, compared with Search-R1 baseline.}
  \label{fig:behavior_evolution}
  \vspace{-0.5em}
\end{figure}

\paragraph{Evolution of Search Behavior}
As illustrated in Figure~\ref{fig:behav_initial}, on the 3B model, our DeSA method achieves a significantly lower deficient search rate compared to Search-R1, which is trained solely with an EM reward (6.96 vs. 23.36). Benefiting from this higher-quality search, DeSA also obtains considerably higher recall and EM scores (64.5 vs. 59.5 and 40.33 vs. 36.00, respectively). Figure~\ref{fig:behavior_evolution} explains how our two-stage training affects the performance. After Stage 1 (Search Skill Acquisition), the agent's deficient search rate drops significantly below the Search-R1 baseline (14.60 vs. 23.36), while its recall already surpasses it (62.55 vs. 59.50). It demonstrates that the agent can already do efficient search after Stage 1. Following this, Stage 2 (Outcome Optimization) brings a sharp increase in the EM rate (from 29.2 to 40.3). Concurrently, the agent refines its search strategy from a slightly exhaustive, recall-focused approach to a more precise one, causing the deficient search rate to fall further to 6.96 and the recall to modestly increase to 64.50. This progression validates the effectiveness of our two-stage training methodology.

\subsection{Ablation Study}

\begin{table*}[t]
\centering
\caption{Ablation study results on the Qwen2.5-3B-Instruct model. DeSA with $R_{\text{recall}}$ at Stage 1 serves as the baseline against which component ablations are compared.}
\vspace{-0.5em}
\label{tab:ablation_results}
\resizebox{\textwidth}{!}{%
\begin{tabular}{lcccccccc}
\toprule
\multicolumn{1}{c}{\textbf{Methods}} & \multicolumn{3}{c}{\textbf{General QA}} & \multicolumn{4}{c}{\textbf{Multi-Hop QA}} & \multicolumn{1}{c}{} \\
\cmidrule(lr){2-4} \cmidrule(lr){5-8}
 & NQ$^\dagger$ & TriviaQA$^*$ & PopQA$^*$ & HotpotQA$^\dagger$ & 2Wiki$^*$ & Musique$^*$ & Bamboogle$^*$ & \textbf{Avg.} \\
\midrule
\textbf{DeSA (Ours)} & 0.375 & 0.575 & 0.397 & \textbf{0.352} & \textbf{0.363} & \textbf{0.134} & \textbf{0.347} & \textbf{0.363} \\
\midrule
\multicolumn{9}{c}{\textit{Stage 1 Reward Design}} \\
\midrule
w/ $R_{\text{recall}}$+$R_{\text{penalty}}$ & 0.370 & 0.567 & 0.397 & 0.337 & 0.342 & 0.112 & 0.306 & 0.347 \\
w/ $R_{\text{acc}}$ & 0.374 & \textbf{0.577} & \textbf{0.404} & 0.347 & 0.346 & 0.125 & 0.306 & 0.354 \\
\midrule
\multicolumn{9}{c}{\textit{Single-stage vs. Two-stage Training}} \\
\midrule
Single-stage ($R_{\text{recall}} + R_{\text{EM}}$) & \textbf{0.386} & 0.563 & 0.383 & 0.348 & 0.330 & 0.132 & 0.307 & 0.350 \\
\bottomrule
\end{tabular}
} 
\vspace{-5pt}
\end{table*}

\subsubsection{Reward Design for Stage 1}
\label{sec:reward design}
We conduct experiments to analyze the impact of different reward formulations in Stage 1, using our primary DeSA model (trained with only $R_{\text{recall}}$) as the baseline for comparison. 
 First, we investigate a composite reward, $R_{\text{recall}} + R_{\text{penalty}}$, which adds a penalty of -0.2 for each rollout if it includes a deficient search behavior (\eg, duplicate or invalid queries). Second, we evaluate a more fine-grained signal, Retrieval Accuracy ($R_{\text{acc}}$), which we define as the proportion of retrieved documents containing a ground-truth answer. As shown in Table~\ref{tab:ablation_results}, manually adding the behavior penalty decreases final performance; while it encourages more stable behavior (more details in Appendix~\ref {app:reward design}), it also appears to restrict the agent’s ability to maximize recall. The fine-grained Retrieval Accuracy reward improves performance on General QA tasks but decreases on complex Multi-Hop QA, resulting in a lower overall score.

\subsubsection{Single-stage vs. Two-stage Training}
To validate our central hypothesis that decoupling search and answering is beneficial, we compare our proposed two-stage method against a single-stage baseline. The baseline is trained from the initial model using a reward function that linearly combines the recall and EM signals ($R = 0.5 \times R_{\text{recall}} + 0.5 \times R_{\text{EM}}$). As shown in Table~\ref{tab:ablation_results}, our two-stage approach outperforms the single-stage variant. This suggests that mixing the search reward (recall) and outcome reward (EM) creates confusing optimization pressures, making it difficult for the agent to learn effectively. In contrast, our decoupled pipeline allows the agent to first master the search process before focusing on answer generation, leading to a more robust agent policy.

\subsubsection{Transition Point of the Two-stage Training}
\label{sec:transition}

\begin{figure}[t]
\centering
\begin{subfigure}[b]{0.48\textwidth}
\centering
    \includegraphics[width=\linewidth]{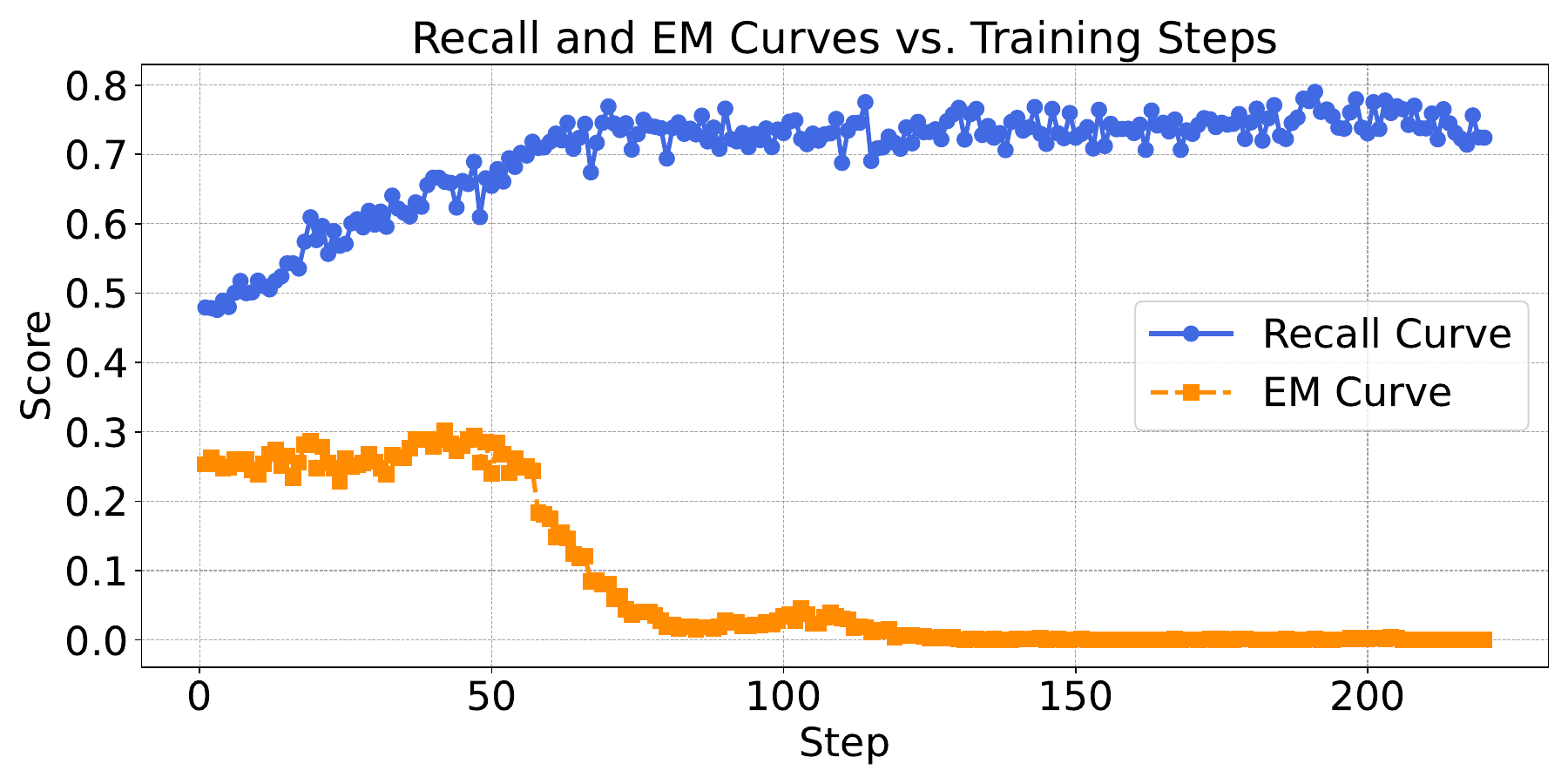}
  \caption{Stage 1 Recall and EM curves.}
    \label{fig:recallandem}
  \end{subfigure}
  \hfill
  \begin{subfigure}[b]{0.48\textwidth}
    \centering
    \includegraphics[width=\linewidth]{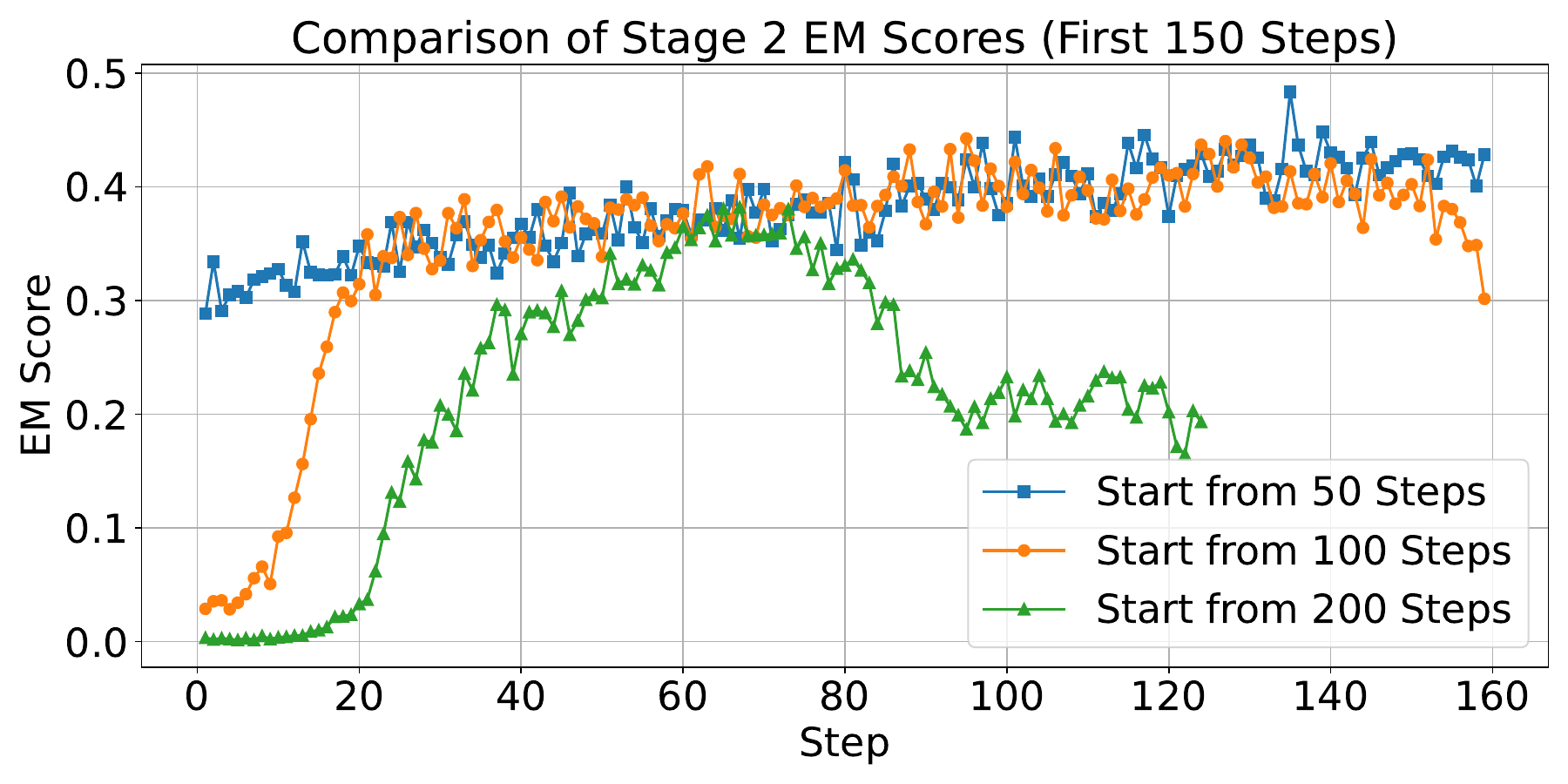}
    \caption{Stage 2 EM curves from different checkpoints.}
    \label{fig:transition}
  \end{subfigure}
  \caption{\textbf{Analysis of the optimal transition point from Stage 1 to Stage 2.}}
  \label{fig:ablation_transition}
\end{figure}

The effectiveness of DeSA depends on selecting an appropriate transition point from Stage 1 to Stage 2. As illustrated in Figure~\ref{fig:recallandem}, continued training in Stage 1 produces a characteristic pattern: the exact-match (EM) score rises to an early peak (around 50 steps in our setting) and then drops abruptly, even though the recall metric continues to improve. This divergence indicates that the model has started to exploit the recall-based reward at the expense of generating correct final answers. 
We therefore use the last pre-collapse checkpoint (the point just before the EM curve turns downward) as a practical marker for transitioning. To test this guideline, we initiate Stage 2 training from checkpoints at 50, 100, and 200 steps. 
As shown in Figure~\ref{fig:transition}, we observe that when Stage 2 begins after the collapse has already started, the agent requires substantially more training to recover its question‑answering behavior, and the second stage becomes less stable. 
This demonstrates that identifying the transition point via the EM curve provides a concrete and effective criterion for moving between stages.

\section{Conclusion}

In our work, we identify and analyze the inefficacy of using outcome-only rewards to optimize the search behavior of agents. We propose DeSA (Decoupling Search-and-Answering), a two-stage training pipeline that separates search skill acquisition from answer generation optimization. Comprehensive experimental evidence demonstrates that DeSA achieves superior search quality (in terms of higher recall and fewer deficient behaviors) and significantly improves performance on QA datasets compared to baselines. Our work represents a re-examination of the current agentic RL training paradigm, emphasizing the value of process-based rewards.

For future work, we plan to explore more advanced process-based rewards for Stage 1 training, such as using a dedicated reward model to evaluate the agent's search behavior. Furthermore, we aim to extend the principle of our DeSA framework to broader agentic tasks beyond question-answering. We believe this principle could also be effective in domains such as code generation and long-context understanding, within both single-agent and multi-agent settings.

\section*{Acknowledgment}

We thank Bowen Jin from the University of Illinois Urbana-Champaign for constructive discussions on Search-R1 baselines and reproduction.


\bibliography{iclr2026_conference}
\bibliographystyle{iclr2026_conference}

\newpage
\appendix

\section{Training Configuration}

For our GRPO training, we use the same training configuration for the two stages. We set the policy LLM learning rate to \(1e^{-6}\) and sample 5 responses per prompt for advantage estimation. The model is trained for a total of 1005 steps, with a learning rate warm-up ratio of 0.285.

Training is conducted on a single node with 4 GPUs. We use a total training batch size of 512. For the policy update, the mini-batch size is 256 and the micro-batch size is 32. The maximum sequence length is set to 4,096 tokens, with a maximum response length of 500 and a maximum length
of 500 tokens for retrieved content. To optimize memory usage, we enable gradient checkpointing and use Fully Sharded Data Parallel (FSDP) with CPU offloading for parameters, gradients, and the optimizer state.

For efficient rollouts, we use the vLLM engine with a sampling temperature of 1.0. The KL divergence regularization coefficient is set to 0.001. During interactions, the maximum number of turns is set to 4, and we retrieve the top 3 passages for each search query. Model checkpoints are saved every 50 steps, and the final checkpoint before collapse is used for evaluation.

\section{Supplementary Results for Stage 1's Reward Design}
\label{app:reward design}

\begin{figure}[htbp]

  \includegraphics[width=0.95\linewidth]{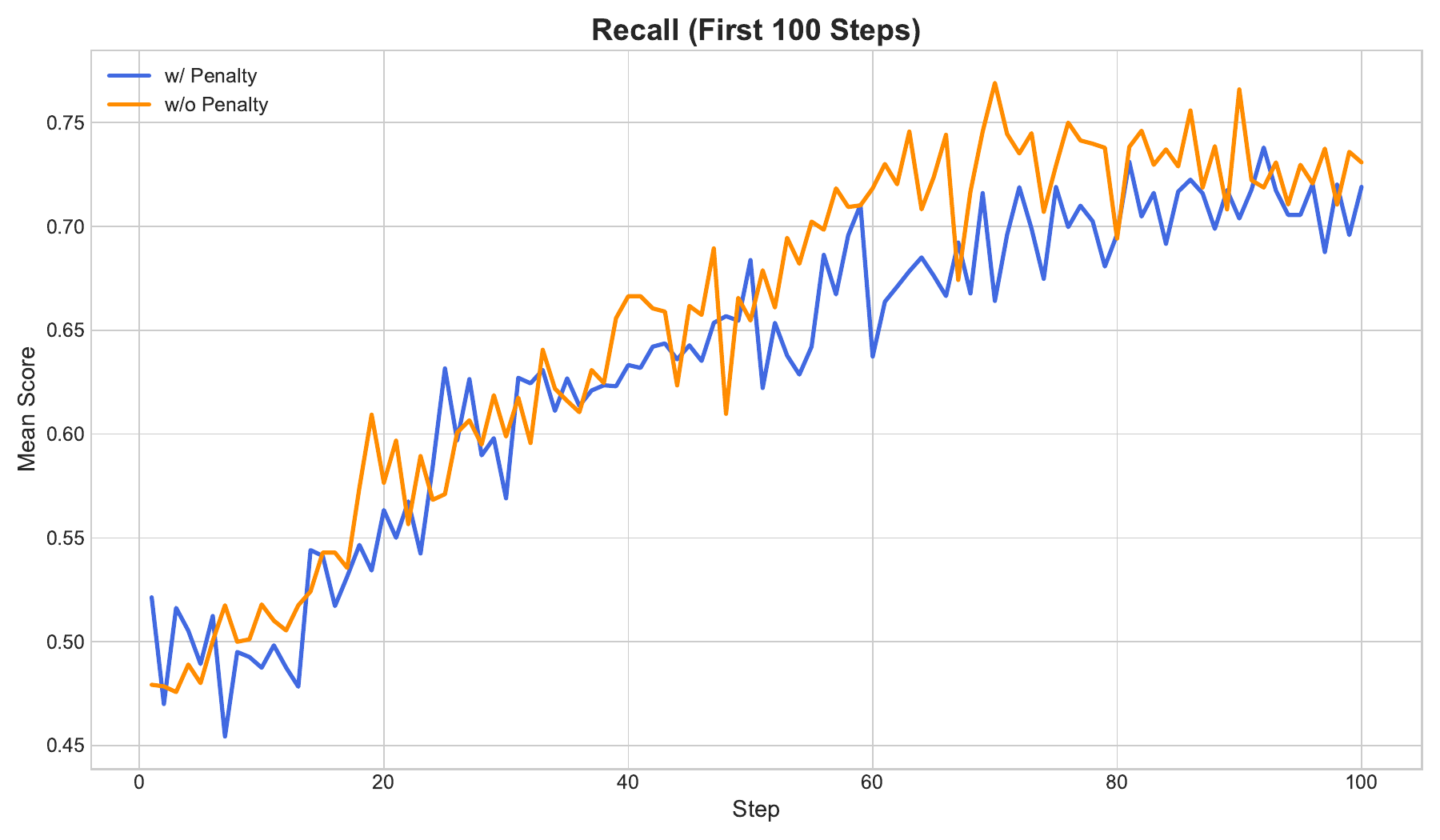}

  \caption{\textbf{Recall Comparison of First Stage.} The blue line is trained with $R_{\text{recall}}+R_{\text{penalty}}$ and the orange line is trained with $R_{\text{recall}}$ only.}
  \label{fig:with_out_penalty}
\end{figure}

Compared to Stage-1 training with $R_{\text{recall}}$ only, $R_{\text{recall}} + R_{\text{penalty}}$ results in a lower deficient search rate (5.09\% vs 14.6\%) on the evaluation sets. However, as shown in Figure~\ref{fig:with_out_penalty}, after about 40 steps, its mean training recall lags behind $R_{\text{recall}}$, which suggests that this somehow restricts the agent from developing its search efficiency. 


\end{document}

%% file: math_commands.tex
\usepackage{amsmath,amsfonts,bm}

\def\eqref#1{equation~\ref{#1}}

\def\1{\bm{1}}

\DeclareMathAlphabet{\mathsfit}{\encodingdefault}{\sfdefault}{m}{sl}
\SetMathAlphabet{\mathsfit}{bold}{\encodingdefault}{\sfdefault}{bx}{n}

